# DeltaVisionX: A Low-Cost Vision-Guided Automation System Using YOLOv11 and PLC-Based Actuation

Lara Sobh[1], Marc Nahhas[1], Francisco Katach El Halabi[1], Michel Owayjan[1, 2], Fadi Elias Sleiman[3]

[1]Faculty of Engineering and Computer Science, American University of Science & Technology, Beirut, Lebanon

[2]Institut de Recherche en Energie Electric de Nantes Atlantique (IREENA), Nantes University, Saint-Nazaire, France

[3]Raymond Feghaly Co., Roumieh, Lebanon

lms20014@students.aust.edu.lb, mmn20013@students.aust.edu.lb, fsk20008@students.aust.edu.lb, mowayjan@aust.edu.lb, F.sleiman@rtf-co.com

*Abstract*—**This paper presents DeltaVisionX, a low-cost industrial automation system that integrates real-time computer vision with PLC-based actuation for object classification and sorting. The system uses the YOLOv11 (You Only Look Once) deep learning model to detect objects on an operating conveyor belt and communicates detection results to a Delta DVP-SE PLC, which actuates pneumatic pistons for sorting. A custom dataset was created, labeled, and augmented to train the YOLOv11 model, achieving a mean average precision of 0.97 with real-time inference of 100 ms per frame. Encoder-based tracking ensures reliable synchronization between detection and actuation, supporting a throughput of 80 objects per minute. Results show that affordable, locally developed automation can reduce labor dependency and improve manufacturing efficiency. The novelty of the proposed system lies in its encoder-synchronized vision–PLC integration rather than in the detection model itself. Future work includes integrating a high-speed Delta robot with EtherCAT-based servo control for advanced pick-and-place operations.**

## I. Introduction

Industrial automation has become a cornerstone for improving operational efficiency, accuracy, and productivity in manufacturing processes worldwide. While highly automated systems are widely adopted in developed markets, many industries in Lebanon continue to rely on manual labor for repetitive tasks such as pick-and-place operations [1]. This reliance often results in inconsistent performance, reduced efficiency, and higher operational costs, limiting the competitiveness of local manufacturers in both regional and global markets.

Despite the clear benefits of automation, existing solutions such as robotic pick-and-place systems or vision-guided manipulators are often expensive and technically complex, making them inaccessible for small and medium-sized enterprises (SMEs). Therefore, there is a critical need for cost-effective, locally implementable automation solutions that can improve productivity while remaining financially viable for Lebanese industries.

This work addresses this gap by designing a piston-based automation system integrated with a computer vision algorithm (YOLOv11) to perform object-pushing tasks on a conveyor belt. The system is intended as a first step toward fully automated sorting and pick-and-place operations, providing a scalable platform that could later incorporate more advanced robotic mechanisms, such as delta robots. By synchronizing the conveyor motion, piston actuation, and real-time object detection, the system aims to enhance operational speed and accuracy compared to manual handling.

The significance of this research lies in providing SMEs with an affordable and effective automation solution that supports productivity, workforce skill development, and technological advancement. In doing so, it contributes to narrowing the automation gap in developing industrial markets.

The study investigates the following research questions:

1) Can the YOLOv11 algorithm achieve real-time object detection with accuracy levels suitable for industrial applications?

2) How effectively can a piston-based mechanism execute precise object-pushing tasks guided by the vision system?

3) What improvements in speed, accuracy, and cost can this system achieve compared to traditional manual sorting methods?

By addressing these questions, this work aims to demonstrate the feasibility and benefits of combining computer vision with simple mechanical automation to enhance industrial productivity in Lebanon and similar contexts.

While previous studies have independently addressed real-time object detection, PLC-based actuation, and high-speed robotic manipulation, their integration into a cost-constrained industrial automation pipeline remains insufficiently explored. Most existing solutions either rely on high-end robotic systems, GPU-based embedded platforms, or tightly coupled motion controllers that are financially and technically inaccessible to small and medium-sized enterprises (SMEs).

Furthermore, state-of-the-art approaches typically assume time-based synchronization or deterministic robotic motion, whereas practical conveyor-based systems are subject to speed fluctuations, mechanical tolerances, and communication latency. These challenges are rarely addressed at the system-integration level.

This work addresses this gap by proposing a low-cost, vision-guided automation architecture that emphasizes encoder-based synchronization between computer vision and PLC-controlled actuation. The primary contribution is not a new detection algorithm, but a practical and scalable integration strategy that enables reliable real-time sorting using standard industrial components under realistic operational constraints.

## II. Literature Review

Computer vision is central to this project, where real-time detection speed and acceptable accuracy are required for industrial automation. Boukabous and Azizi categorize object-detection algorithms into one-stage and two-stage approaches, noting that single-stage detectors like YOLO sacrifice minimal accuracy for substantially faster performance, making them more suitable for real-time video analysis [2]. Redmon et al. introduced YOLO as a unified detector that frames object detection as a regression problem, predicting bounding boxes and class probabilities in a single network pass and achieving real-time frame rates that outperform region-based methods like DPM and R-CNN [3]. Do et al. applied YOLO with image processing for delta-robot product sorting, demonstrating 97.8% accuracy in object identification and real-time control via PLC [4].

Several studies validate YOLO's effectiveness in industrial and embedded scenarios. Anushkannan et al. used YOLO for safety helmet detection in industrial environments [5], while Wu and Wang applied YOLOv7 for traffic sign detection on embedded hardware, showing that lightweight YOLO variants run efficiently on resource-limited platforms [6]. Jamtsho et al. reported high accuracy for license-plate localization using YOLOv2 [7], and Yarmohammadi et al. integrated YOLO-based detection with a delta robot for chess piece manipulation, achieving 90% mAP and 95% grasping success rate [8]. Li et al. compared YOLOv3, Faster R-CNN, and SSD for greenhouse detection in satellite imagery, finding that YOLOv3 achieved the highest mAP (90.4%) and substantially higher FPS (73) than Faster R-CNN (12 FPS) and SSD (35 FPS), confirming YOLO's advantage for real-time high-resolution applications [9]. More recent studies on YOLOv10 and YOLOv11 highlight architectural improvements that eliminate Non-Maximum Suppression (NMS) and further reduce inference latency, enabling faster deployment on edge devices [10-11]. Mao et al. demonstrated that YOLOv10 improved accuracy (85.4%) and speed for fabric defect detection over previous YOLO variants [13], while Reshetnikov and Ronkin used YOLOv8 with oriented bounding boxes to estimate rock fragment sizes on conveyor belts, illustrating YOLO's suitability for irregular industrial objects [12].

While YOLO dominates recent real-time detection work, alternative detectors have also been explored for high-accuracy applications. Fan et al. showed that Faster R-CNN achieved 89.06% AP in vehicle detection by optimizing anchor sizes and learning rates, demonstrating superior performance in scenarios with overlapping objects where YOLO can struggle [13]. Sun and Wu enhanced the SSD algorithm with Batch Norm to improve generalization and built a detection system using Flask for real-time result display [16]. These comparisons underscore YOLO's speed advantage for embedded and real-time systems, justifying its selection for this project.Recent comparative studies have further evaluated the performance of emerging YOLO variants. Harminto et al. compared YOLOv5, YOLOv8, and YOLOv11 for image segmentation tasks, highlighting trade-offs between accuracy and computational complexity [14]. Similarly, Gorro et al. demonstrated the effectiveness of newer YOLO architectures, for instance segmentation under challenging visual conditions [15]. These studies confirm the suitability of modern YOLO variants for real-time industrial vision tasks, while reinforcing the need for efficient system-level integration.

In addition to vision algorithms, real-time communication protocols are critical for integrating detection outputs with industrial controllers. Nguyen et al. demonstrated that EtherCAT provides deterministic, high-speed communication with sub-millisecond cycle times for multi-axis motion control and automated guided vehicles [16]. Zhang et al. and Kelkar designed EtherCAT master systems on embedded platforms to solve bandwidth and synchronization challenges in multi-axis motion control [17]. Hung et al. used EtherCAT between a robotic arm and PC (with C#) to achieve precise synchronous control of each axis [18], while Hu and Xu designed a real-time control system for a multi-locomotion mobile robot using EtherCAT with LabWindows and CODESYS software [19]. Ahn et al. resolved latency issues by implementing a dual-channel EtherCAT configuration, enhancing communication efficiency in a 33-DOF humanoid robot [20], and Kim et al. used EtherCAT in a parallel robot system with Windows RTX for real-time computation and motion control [21]. In contrast, Kaewkorn et al. and Michalík et al. used TCP/IP-based protocols for communication between PLCs and vision processing nodes (including Raspberry Pi) in delta robot systems, demonstrating that Ethernet-based links can maintain acceptable latency (960 μs to 71 μs) while simplifying integration with industrial equipment [22].

Delta robots are widely used for high-speed pick-and-place operations due to their parallel kinematic structure, which minimizes inertia and enables rapid motion with high precision. Gasparetto and Scalera trace the development of delta robots from early parallel manipulators through commercial success in packaging industries, highlighting their capability for up to 200 pick-and-place operations per minute [23]. Hadfield et al. analyzed forward and inverse kinematics using Conformal Geometric Algebra, emphasizing the importance of singularity avoidance and real-time control algorithms for stable performance [24]. Several studies have integrated delta robots with vision systems: Paudel et al. combined a delta arm with YOLOv8 for automated waste sorting, achieving real-time classification but facing challenges in coordinating image processing with robotic motion [25]; Zhang developed a CODESYS-based control system for vision-guided random bin picking, demonstrating the feasibility of combining delta robots with vision sensors [26]; and Cheng et al. proposed a systematic hand-eye calibration framework for delta robot pick-and-place systems, achieving fast calibration within 10 minutes [27]. Öztürk and Kuncan integrated a linear delta robot with a PLC using image processing for pick-and-place operations [28]. Similarly, Mojtahedi et al. carried out an experimental study on autonomous food packaging using a delta parallel robot with a two-fingered gripper, highlighting the suitability of delta architectures for high-speed pick-and-place tasks in packaging lines [29].

## III. System Architecture and Engineering Methodology

The problem addressed in this work is primarily an engineering and system-integration challenge rather than a purely algorithmic research problem. To implement the proposed system, hardware and software components were selected based on cost-effectiveness, availability, technical reliability, and compatibility with industrial devices such as actuators and conveyors. The critical components that determine the system's performance are discussed below.

For the processing unit, several alternatives were considered. These included a Raspberry Pi, which supports Python and Linux-based applications; an Arduino Mega, suitable for basic microcontroller tasks but lacking native Python support; and an NVIDIA Jetson Nano, which enables GPU-accelerated AI and computer vision workloads. For sensing, standard cameras, depth sensors, and high-speed industrial cameras were evaluated to determine the most effective solution for object detection and real-time tracking.

Regarding software, Python, C++, and MATLAB were examined as potential programming environments due to their suitability for computer vision applications and integration with hardware platforms.

The final design employs the YOLOv11 detection model running on a standard laptop, which communicates with a Delta DVP-SE PLC for actuator control. This configuration was selected for its balance of technical feasibility, cost efficiency, and development simplicity. The laptop provides sufficient processing power for real-time object detection without the need for specialized hardware, while the PLC ensures reliable control of actuators. Familiar programming languages and widely available hardware components were used to minimize development time and facilitate system integration (Fig. 1).

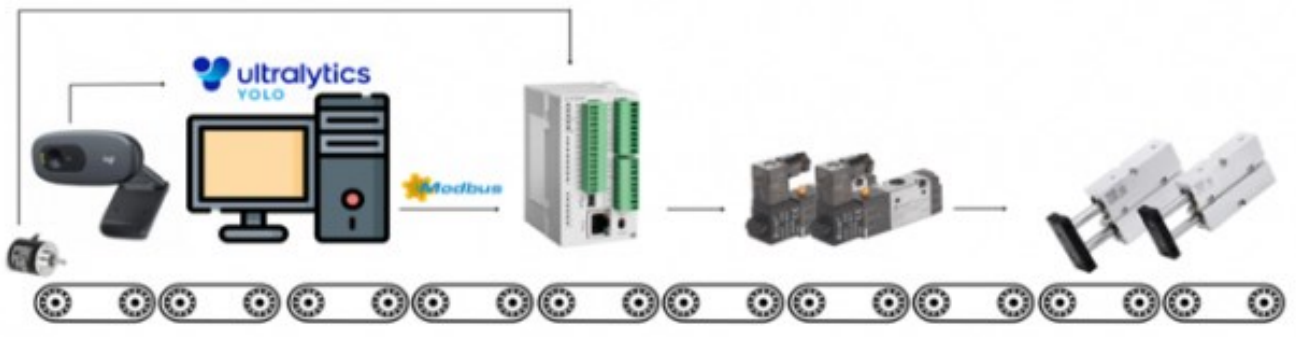

Fig. 1. Selected Design Approach

### A. *Design Specifications*

The system was designed to classify and sort objects on a conveyor at a rate of 80 items per minute, with each image processed in under 100 milliseconds. The physical frame was limited to a tabletop size, approximately 1 meter long. A 220 V AC supply was used to power the conveyor, a 12 V DC supply was used to power the actuators, while the image processing unit operated on a 5 V input. The camera resolution was set to 640x480 pixels, the region of interest is a rectangle inside that frame, and all components were configured to operate in standard indoor lighting conditions.

### B. *Multiple Realistic Design Constraints*

Several constraints were encountered during development. The overall project budget was deliberately limited to align with the needs of small businesses, so low-cost materials were prioritized. Due to time constraints for training on Delta robots, the initial setup relied on piston actuators. Power limitations also influenced the selection of computing hardware, excluding more powerful but energy-intensive options. Additionally, the image processing model had to maintain consistent performance despite varying ambient lighting conditions and fluctuating conveyor speeds, which added complexity to the image processing model.

### C. *Materials*

The project materials included a laptop running Windows and Python for model inference and control logic, a USB webcam for image capture, a Delta DVP-SE PLC for actuator control, and pneumatic pistons for object sorting. The PLC was programmed using ladder logic via ISPSoft software. Python libraries such as Ultralytics YOLO and Roboflow APIs were used for dataset management and model training. These materials were chosen based on cost, availability, ease of integration, and suitability for the task, which makes it a practical and scalable solution for local businesses.

### D. *Methods*

The system was designed with three main components: image acquisition, object detection and classification, and actuator control. Each component was carefully implemented to ensure reliable and accurate sorting of chocolate bars on a moving conveyor. Actuator selection, piston speed, and communication protocol choice are treated as design constraints rather than research contributions, and were selected based on availability, cost, and industrial compatibility.

In the image acquisition component, a webcam was positioned above the conveyor belt to continuously capture frames. Within each frame, a specific region of interest (ROI) was defined. Detection and tracking of objects were limited to this ROI to avoid multiple counting of the same object.

For the object detection and classification component, the laptop processed the acquired images using a trained YOLOv11 model. The model was designed to classify detected objects into one of four predefined classes, corresponding to different types of chocolate bars labeled as classes 0, 1, 2, and 3. Each classification triggered a unique signal to be sent to the PLC. The class number identified by YOLO was mapped directly to the activation of the corresponding actuator, so that the system differentiates the handling of each chocolate type based on detection results.

The actuator control component follows a queuing system, which was essential for precise timing and sorting on the conveyor belt. Each type of chocolate bar had its own dedicated memory stack within the PLC. This memory stack stored the positions of detected objects based on encoder pulses rather than absolute position values. The encoder was mounted beneath the conveyor and produced pulses corresponding to the exact movement of the belt, such that variations in conveyor speed did not affect positional tracking accuracy.

At the moment YOLO detected an object, the current encoder pulse count was immediately saved to the appropriate memory stack in the PLC, corresponding to the chocolate bar's class. The PLC continuously monitored the encoder in real time, calculating the difference between the current encoder pulse count and the stored initial pulse count for each detected object. This difference was compared to the predefined distance from the camera's perpendicular projection point to the corresponding actuator piston. When the encoder pulse count difference reached this target distance, the PLC activated the associated piston to push the chocolate bar into its designated bin.

The memory stack of each piston functioned as a queue; it allowed multiple chocolate bars of the same type to be tracked simultaneously. This was necessary because multiple bars frequently appeared on the conveyor with short gaps between them. By using stacks (or cyclic buffers), the system maintained the order of objects, ensuring that each chocolate bar was pushed at the correct time and that no objects were missed or processed out of sequence (Fig. 2).

When an object is processed and about to be pushed, the pointer in the corresponding memory stack increments to the next cell. If the pointer reached the end of the stack, it wrapped back to the first cell, overwriting the oldest data in a continuous loop. This cyclic memory management ensured that memory usage remained dynamic and efficient, avoiding overflow issues and reducing the required memory footprint in the PLC. Such a design made the system capable of handling continuous and variable flow of objects without degradation in performance.

By combining encoder-based position tracking with class-specific memory stacks and synchronized piston activation, the actuator control system achieved high precision and reliability. Using encoder pulses to represent position, instead of relying on time or speed estimates, was crucial in eliminating sorting errors caused by conveyor speed variations. However, a limitation remains when two chocolate bars are placed very close together or stacked, which can cause detection and sorting issues. This limitation would be resolved if a Delta robot replaced the pistons, as the system would then track and use object coordinates in the queue for more precise handling.

## IV. Results and Discussion

### A. *Testing and Validation*

Software testing was performed by generating multiple datasets during development to ensure that training was properly conducted. These datasets mainly included images of various candies and chocolate bars.

Creating each dataset involved capturing photos of the selected products from different angles, under different lighting conditions, and at different speeds (both while stationary and while moving on the conveyor to simulate real-time operation). After collecting around 100 images, they were manually labeled using Roboflow (Fig. 3). This involved drawing bounding boxes around each object. For datasets used with YOLO-OBB (YOLO with oriented bounding boxes), polygons with multiple vertices were drawn to accurately outline each object.

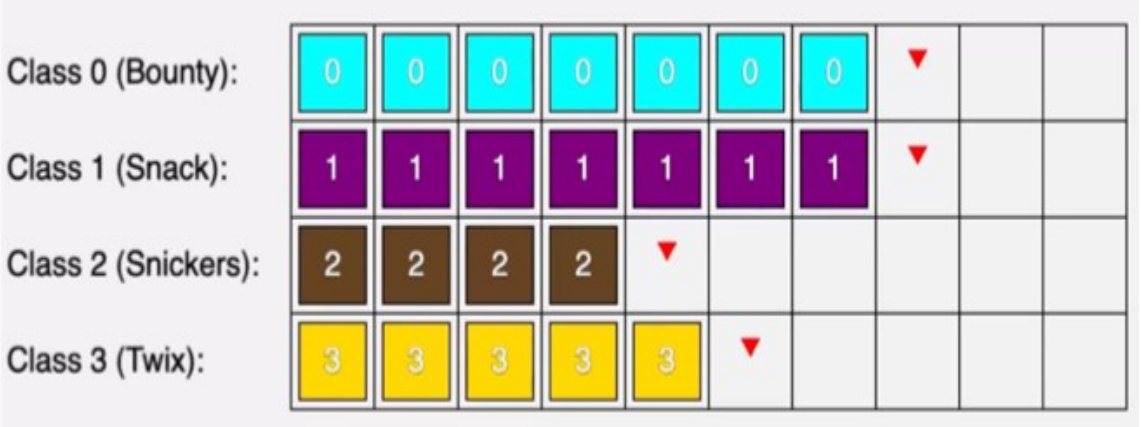

Fig. 2. Simulating the queuing system

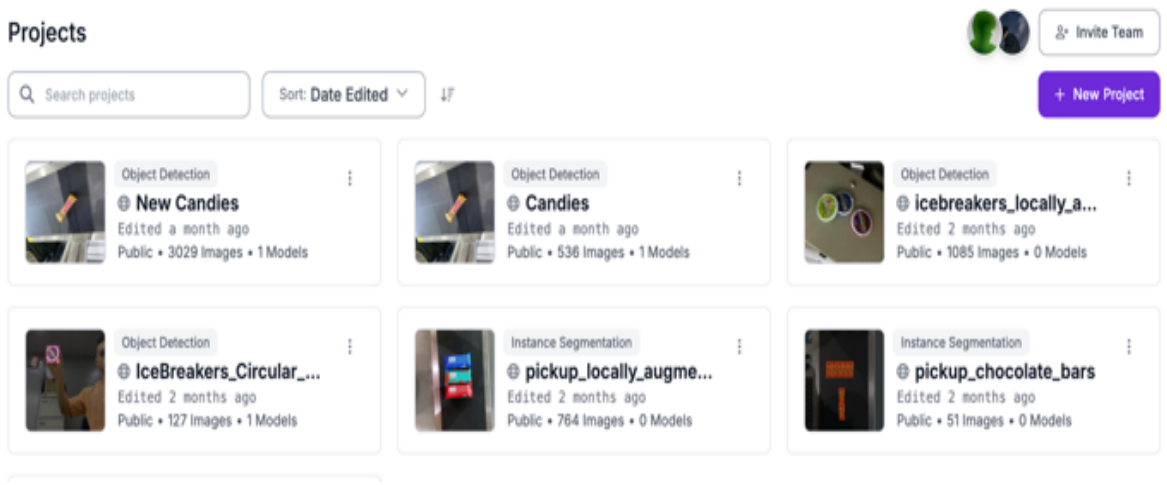

Fig. 3. Creation of multiple datasets in Roboflow

Each dataset then underwent a second round of augmentation using a custom local script. Augmentation, which is the process of artificially expanding a dataset by modifying existing images, included transformations such as rotation, noise addition, and mosaic effects, while also adjusting the corresponding labels accordingly.

The next step was model training, which was done on Google Colab. This began by installing the required libraries, including Ultralytics and Roboflow. An API key from Roboflow was used to access a specific project and version, which provided the labeled dataset in a format compatible with YOLOv11. This API key is unique and allows secure access to the dataset.

A lightweight YOLOv11 model (yolo11n.pt) was then imported from the official Ultralytics release. Training was set to run for 500 epochs, with early stopping (patience) of 50 epochs, which means training would automatically stop if the performance of the model did not improve for 50 consecutive epochs. After training, key outputs such as performance metrics and a confusion matrix were displayed to evaluate the model's accuracy and effectiveness. Although Colab uses a T4 GPU, each training session still required several hours to complete. Fig. 4 depicts the confusion matrix and testing/validation metrics.

Hardware testing began with the PLC (DVP-SE), which was verified using a very simple setup with lamps. Each lamp (1, 2, or 3) was activated depending on the type of candy detected (Fig. 5).

The conveyor belt was also tested at various speed levels to assess the stability of object detection and system synchronization under different conditions. Communication between the PLC and the Python program running YOLO detection was established via TCP/IP. The PLC was programmed using ISPSoft with ladder diagram, and the monitor table feature was used to visualize detections in real time, showing which lamp was turned on for each detected object.

The same example was later reproduced using pistons instead of lamps (Fig. 6). For future upgrades involving the Delta robot, similar testing can be performed using the same monitor table to display the live coordinates of objects detected on the conveyor.

### B. *Results*

After training the YOLO model on Google Colab, several output metrics were generated to evaluate its performance.

On the hardware side, indicator lamps and pistons responded consistently and accurately according to detected object classes. All results confirmed that the detection system functioned reliably under different lighting, angles, and conveyor speeds, with an achieved rate of 80 objects per minute and approximately 100 ms image processing time.

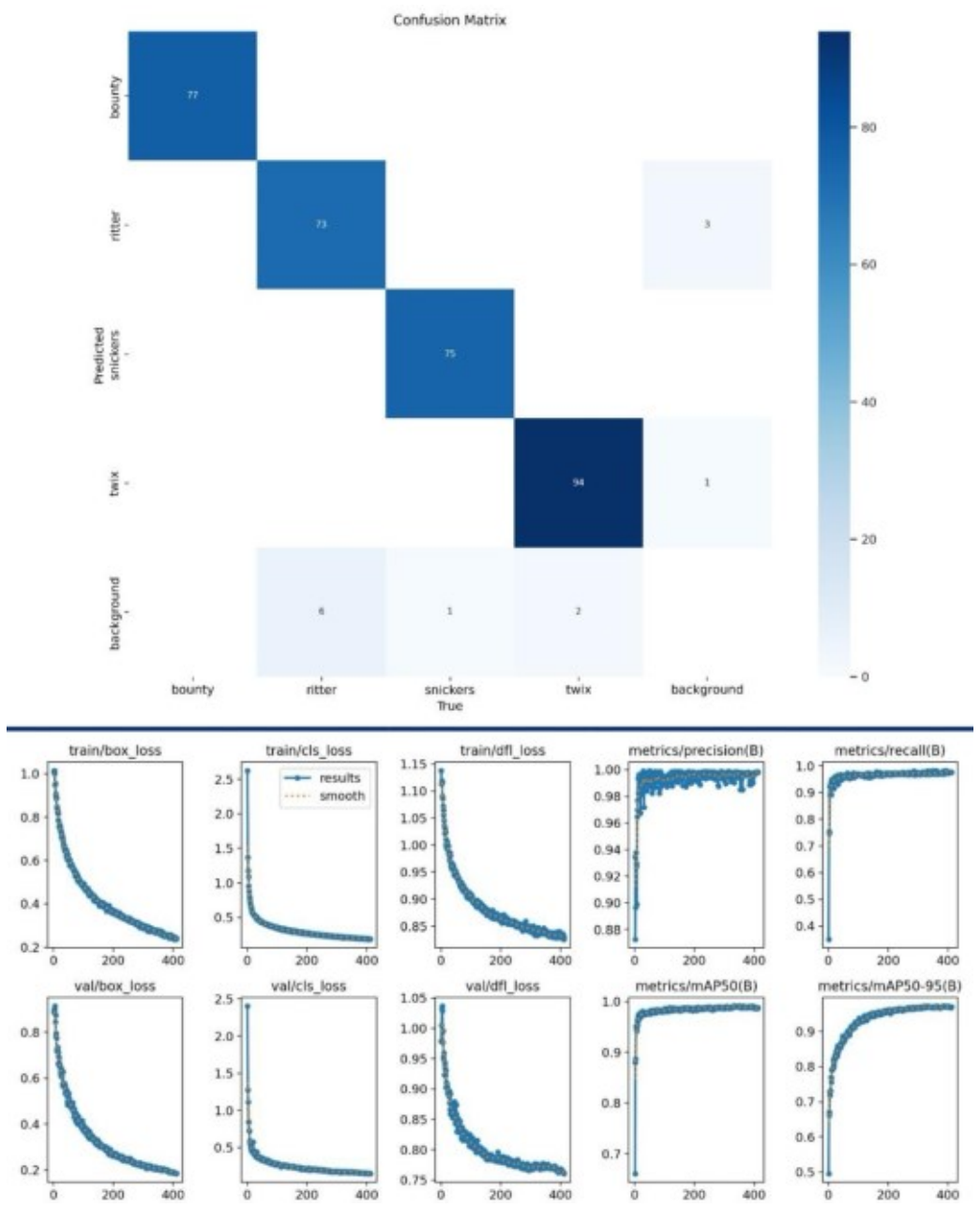

Fig. 4. Confusion matrix and training/validation metrics

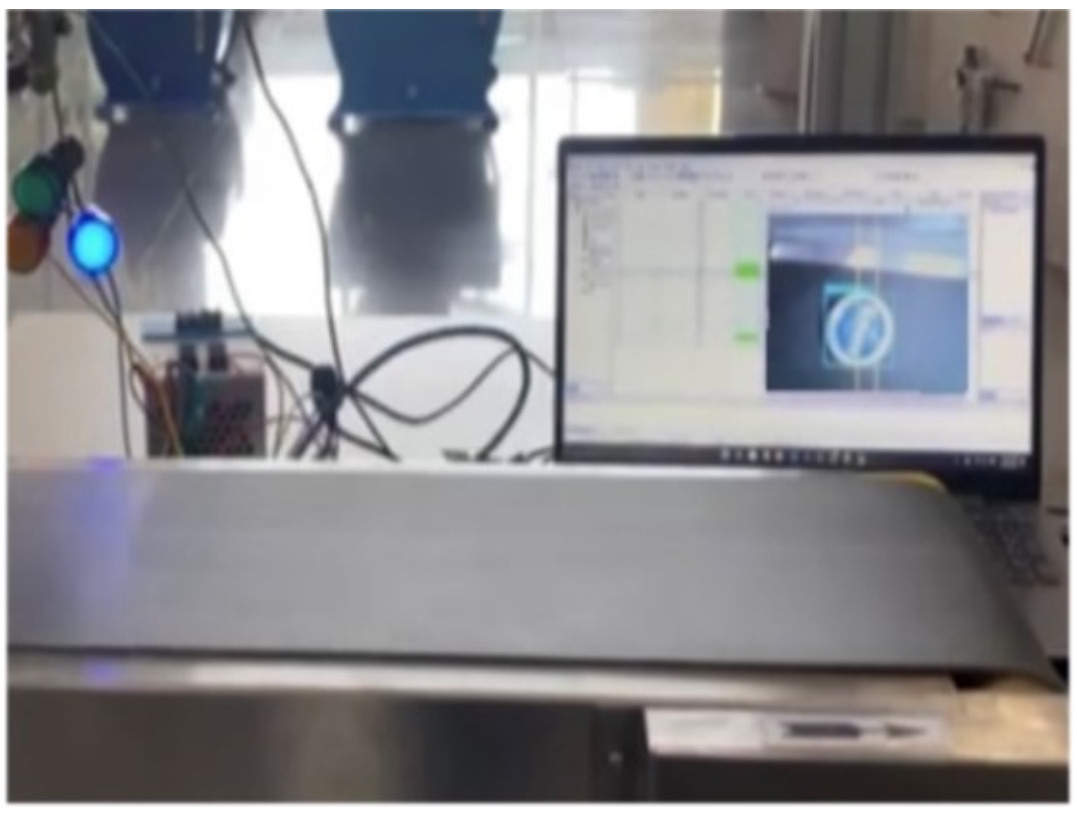

Fig. 5. Initial demo setup for testing

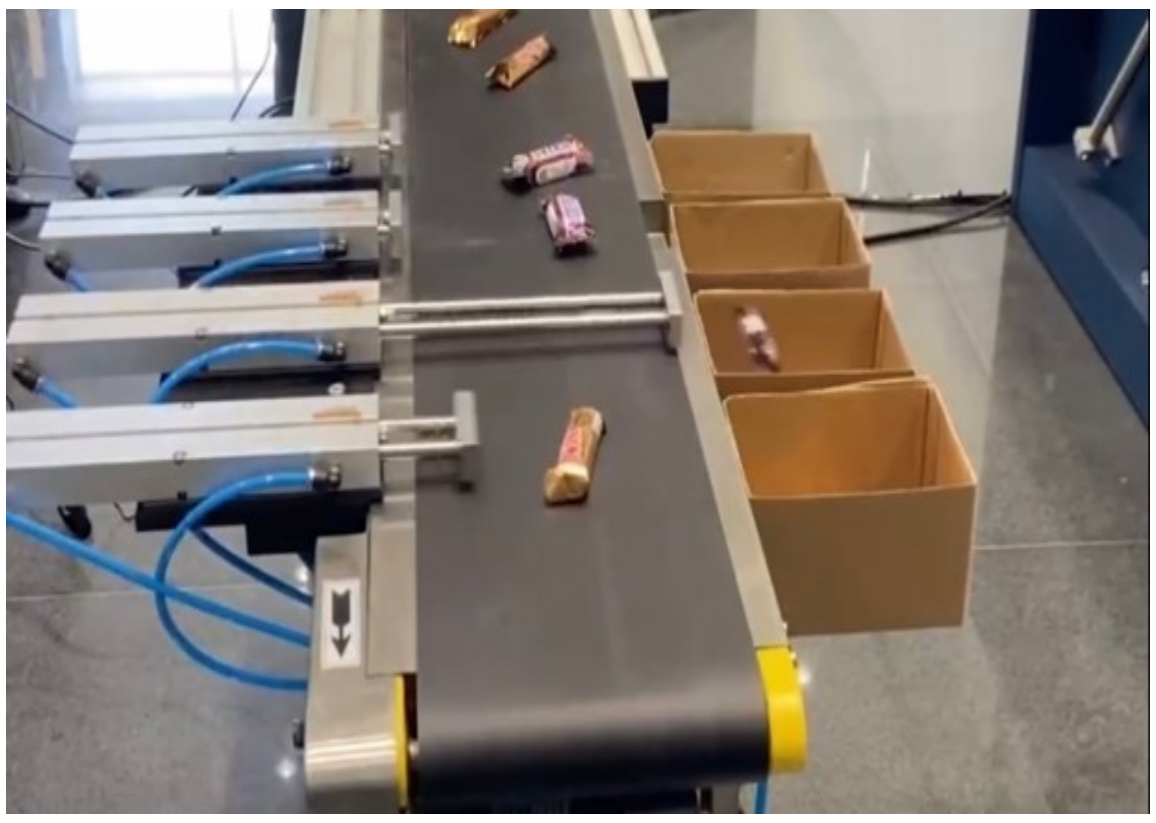

Fig. 6. Final demo setup

### C. Discussion

The development and testing of the system provided valuable insights into both the strengths and limitations of integrating YOLO-based object detection with low-cost automation components. One of the most important lessons learned was the importance of clean, well-prepared datasets. Creating multiple datasets throughout the project not only supported training but also helped build experience in data collection, labeling, and augmentation. This was crucial for improving detection accuracy and minimizing inconsistencies.

However, several limitations were encountered during testing. A recurring issue was false positives—objects being incorrectly classified, such as a transparent water bottle being detected as a mint candy. After analysis, it was suspected that color similarity between objects contributed to this confusion. A practical solution for future work would be to refine dataset labeling and ensure the training images include a wide variety of similar-looking non-target objects to help the model learn finer distinctions.

The system showed that even when paired with simple actuators like pistons, the YOLO model was fast enough to process images and return results in real time. This confirms that the model is more than capable of handling faster and more complex tasks. Based on this performance, it is realistic to assume that the system would operate even more efficiently when upgraded to a Delta robot setup with servo motors and EtherCAT-based communication. This future configuration would likely allow for higher-speed object handling and more precise pick-and-place actions.

Although individual components of the proposed system – such as object detection models and PLC-based control – are well established, their reliable integration under real-time, low-cost, and non-deterministic conveyor conditions remains an open practical challenge. This work demonstrates that encoder-based synchronization offers a robust solution to this problem, while highlighting remaining challenges related to object proximity, occlusion, and scalability toward robotic manipulation.

In summary, while the current implementation successfully demonstrated real-time classification and actuation using affordable components, it also highlighted areas that require further refinement. The results are encouraging and justify confidence in both the approach and the technology, provided that the identified limitations are addressed with practical adjustments in future iterations

## V. Conclusions and Future Work

### A. Conclusions

The main objective of this project was to create a cost-effective industrial system for object classification using pistons. This was done to support Lebanese industries that still depend largely on manual labor. The system was designed to detect, track, and push objects using a combination of computer vision, motion control, and communication tools. A YOLOv11-based detection model was used to recognize different shapes, sizes, and orientations of objects under varying lighting conditions. High detection accuracy and reliability were achieved after refining the augmented dataset and optimizing the model.

The most critical challenge was to ensure synchronization between the moving conveyor, the vision system, and the piston actuator. This problem was handled using an encoder. One limitation that remained was the handling of objects that enter side by side or stacked on top of each other. This can be solved by placing a bridge at the beginning so that objects enter separately, one by one. Alternatively, Delta robots can be used to pick objects from any position.

Another important outcome was the proof that a local automation system could be built at a much lower cost compared to imported solutions. This makes it suitable for small and medium-sized industries in Lebanon. Economic and industrial feasibility were confirmed, but some challenges remain, such as workforce adaptation, initial setup costs, and training requirements. The reliability of the results was confirmed through testing under real conditions, where consistent detection and actuation were observed. The approach taken in this project combined low-cost components, an optimized deep learning model, and simple motion control strategies. Alternative solutions such as Delta robots and more advanced control systems were studied and planned for future development. These alternatives were found to be more complex but necessary for handling faster or more delicate operations in the future.

### *B. Future Work*

Future work includes mainly replacing the piston with a Delta robot to enable more flexible and precise pick and place operations instead of just pushing objects. The Delta robot will be powered by servo motors, which will be controlled by a motion controller. EtherCAT communication will be used between the motion controller and the motors to ensure fast and real-time control. Further improvements should include simplifying the user interface, reducing setup time, and providing training materials for operators. More testing is also needed at higher conveyor speeds and with different object types to check the limits of detection accuracy and actuator performance. On-spot training using a local GPU and cloud-based monitoring should also be added to improve adaptability and remote supervision. These steps will help create a more complete and reliable automation system that supports local industry with lower costs and higher productivity.

- The below form will not be published, but it is necessary to help with the review process.

# Authors' Background

| Name | Email | Position (Prof , Assoc. Prof. etc.) | Research Field | Homepage URL |
|---|---|---|---|---|
| Lara Sobh | lms20014@students.aust.edu.lb | Student | Automation | |
| Marc Nahhas | mmn20013@students.aust.edu.lb | Student | Automation, AI | |
| Francisco Katach El Halabi | fsk20008@students.aust.edu.lb | Student | Automation, AI | |
| Michel Owayjan | mowayjan@aust.edu.lb | Prof | Control, Automation, ML | |
| Fadi Elias Sleiman | F.sleiman@rtf-co.com | Senior Sales Engineer | Automation, Control | |